\documentclass[journal]{IEEEtran}
\usepackage{orcidlink}
\ifCLASSINFOpdf
\else
   \usepackage[dvips]{graphicx}
\fi
\usepackage{url}
\usepackage{array}
\usepackage{graphicx}

\begin{document}

\title{R3D-SWIN:Use Shifted Window Attention for Single-View 3D Reconstruction}

\author
{Chenghuan Li
\hspace{-1.0mm}$^{~\orcidlink{0009-0006-0363-2240}}$,
Meihua Xiao$^*$
\hspace{-1.0mm}$^{~\orcidlink{0000-0003-2717-7060}}$,
Zehuan li
\hspace{-1.0mm}$^{~\orcidlink{0009-0001-0586-3601}}$,
Fangping Chen,
Shanshan Qiao,
Dingli Wang
\hspace{-1.0mm}$^{~\orcidlink{0009-0004-7545-5111}}$,
Mengxi Gao
\hspace{-1.0mm}$^{~\orcidlink{0009-0001-0044-773X}}$,
Siyi   Zhang

\thanks{This research was funded by the National Natural Science Foundation of China, grant number 62362033, 61962020), Double Thousand Talent Plan of Jiangxi Province (No. jxsq2023201009), Natural Science Foundation of Jiangxi Province (No. 20224ACB202006).
(Corresponding author: meihua xiao.)}
\thanks{Li et al. are with the Software of School, East China Jiaotong  University, Nanchang 330000, China (e-mail: chenghuanli@ecjtu.edu.cn;
meihuaxiao@ecjtu.edu.cn;zehuanli@ecjtu.edu.cn;shanshanqiao
2022@ecjtu.edu.cn;dingliwang@ecjtu.edu.cn;mengxigao@ecjtu.edu.cn;
siyizhang@ecjtu.edu.cn).

Fangping Chen are with the Jiangxi University of Software Professional Technology,Nanchang 330000, China (e-mail: chenghuanli@ecjtu.edu.cn).
}

}

\markboth{}
{Shell \MakeLowercase{\textit{et al.}}: Bare Demo of IEEEtran.cls for IEEE Journals}
\maketitle

\begin{abstract}
Recently, vision transformers have performed well in various computer vision tasks, including voxel 3D reconstruction. However, the windows of the vision transformer are not multi-scale, and there is no connection between the windows, which limits the accuracy of voxel 3D reconstruction. Therefore, we propose a voxel 3D reconstruction network based on shifted window attention. To the best of our knowledge, this is the first work to apply shifted window attention to voxel 3D reconstruction. Experimental results on ShapeNet verify our method achieves SOTA accuracy in single-view reconstruction.
\end{abstract}

\begin{IEEEkeywords}
Deep learning, Single-View 3D Reconstruction, Voxel model
\end{IEEEkeywords}

\IEEEpeerreviewmaketitle

\section{Introduction}

\IEEEPARstart{S}{ingle-view}  3D reconstruction involves restoring the shape of an object based on a single-view image of the object. This is a very challenging research topic that involves the fields of computer vision and computer graphics. One of the main challenges is how to extract features from a single image to generate corresponding 3D objects. Currently, deep learning reconstructors provide three solutions, including RNN-based methods\cite{ref1}\cite{ref2}, CNN-based methods\cite{ref3}\cite{ref4}\cite{ref5} and transformer-based methods\cite{ref6}\cite{ref7}\cite{ref8}\cite{ref9}\cite{ref10}. In this work, we focus on transformer-based method to improve the accuracy and robustness of single-view 3D reconstruction with voxel representation.

The Vision Transformer\cite{ref11} segments the image into a series of fixed-size non-overlapping patches, which are then flattened and fed into the Transformer model as a sequence. Each patch is considered a position in the sequence. Then, through the Transformer’s attention mechanism, the model can learn the relationship between different positions in the image, thereby capturing global contextual information.However, Vision Transformer relies heavily on large datasets, the fixed feature map size loses multi-scale information, and there is no connection between patches. At the same time, using patches as sequences still does not break away from the curse of trainable sequence length in the field of NLP. As a result, the transformer-based 3D reconstruction method\cite{ref6}\cite{ref7}\cite{ref8}\cite{ref9}\cite{ref10} cannot use pre-trained models with resolutions above 224 pixels.

Recently, Swin Transformer\cite{ref12} has shown great promise as it integrates the advantages of both CNN and Transformer. On the one hand, it has the advantage of CNN to process image with large size due to the local attention mechanism. On the other hand, it has the advantage of Transformer to model long-range dependency with the shifted window scheme\cite{ref12}.

In this paper, we propose an 3d reconstruction model, namely R3D-SWIN, based on Swin Transformer. More specifically,R3D-SWIN consists of tow modules: transformer encoder, CNN decoder.Inspired by 3dretr\cite{ref8}, we designed a simple CNN decoder. In contrast to 3dretr\cite{ref8}, we removed the transformer layer in the CNN decoder. The ablation experiment showed that the decoder module continued to perform self-attention too much, leading to overfitting of the model.

The contributions can be summarized as follows:(1 We apply shifted window attention to 3D reconstruction for the first time.(2 We designed a simple CNN decoder.(3 Experimental results on ShapeNet\cite{ref14} demonstrate that our method outperforms other SOTA methods in single-view reconstruction. Additional experiment results on Pix3D\cite{ref15} also verify its effectiveness on realworld data.

\section{Method}
\begin{figure*}
\centerline{\includegraphics[width=\textwidth,height=4cm]{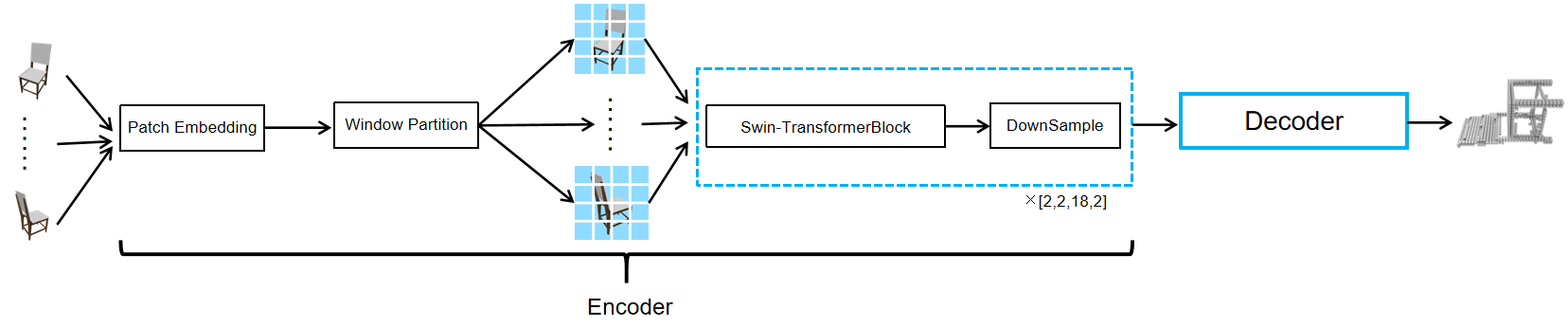}}
\caption{Illustration of our proposed R3D-SWIN and its details.}
\label{fig:model}
\end{figure*}

The overall framework of our proposed method is illustrated in Fig. \ref{fig:model}.The view image set of an object $I= \left \{ I_{1},I_{2},I_{3},\cdots ,I_{N} \right \}$is processed by the encoder $E$ to extract the feature representation for reconstruction. Then the decoder $D$ generates the corresponding voxel-based 3D shape $O$. The overall process is formulated as:
\begin{equation}
O=D\left ( E\left ( I \right ) \right )
\end{equation}

\subsection{Encoder}
\begin{figure}
\centerline{\includegraphics[width=\columnwidth]{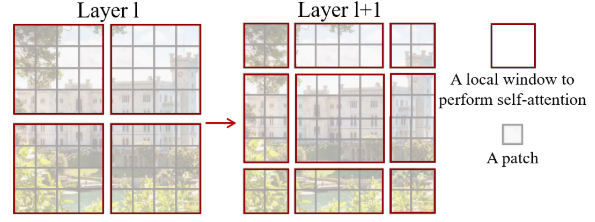}}
\caption{An illustration of the shifted window approach for computing self-attention in the proposed Swin Transformer architecture. In layer l (left), a regular window partitioning scheme is adopted, and self-attention is computed within each window. In the next layer l + 1 (right), the window partitioning is shifted, resulting in new windows. The self-attention computation in the new
windows crosses the boundaries of the previous windows in layer l, providing connections among them\cite{ref12}.}
\label{fig:swintranformerblock}
\end{figure}

The encoder is based on the architecture of Swin Transformer\cite{ref12} and is divided into four stages.Taking Swin-Transformer v2 base as an example,Each stage contains 2, 2, 18, and 2 swin transformer blocks with a downsample respectively. Each swin transformer block contains two layers, as illustrated in Fig. \ref{fig:swintranformerblock}.

\begin{figure}
\centerline{\includegraphics[width=\columnwidth]{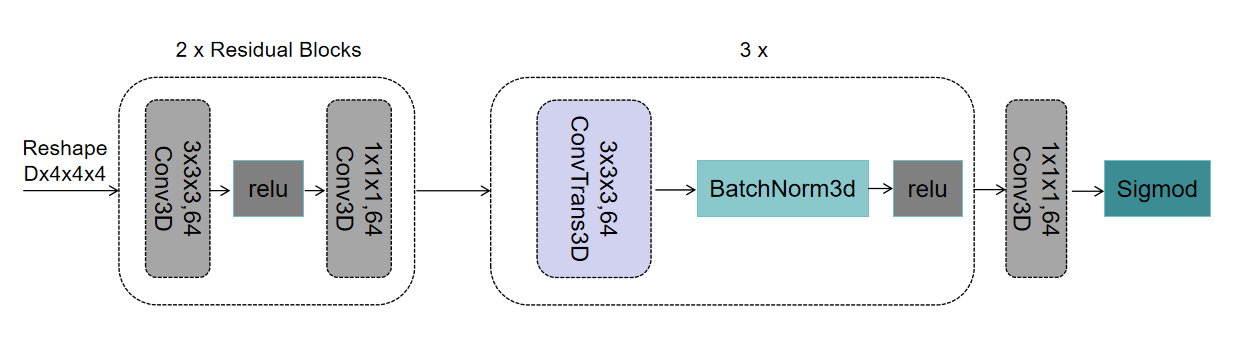}}
\caption{An illustration of the decoder.}
\label{fig:decoder}
\end{figure}

\subsection{Decoder}
Our decoder is similar to 3dretr\cite{ref8},in contrast to 3dretr\cite{ref8}, we removed the transformer layer in the CNN decoder and we insert BatchNorm3d layer after ConvTranspose3d layer as illustrated in Fig. \ref{fig:decoder}.

\subsection{Loss function}
Following 3dretr\cite{ref8}, we use Dice loss\cite{ref16} as the loss function which it is suitable for high unbalanced voxel occupancy. The Dice loss could be formulated as:
\begin{equation}
\iota = 1-\frac{\sum_{i=1}^{32^{3}}p_{i}g_{i}}{\sum_{i=1}^{32^{3}}p_{i}+g_{i}}-\frac{\sum_{i=1}^{32^{3}}(1-p_{i})(1-g_{i})}{\sum_{i=1}^{32^{3}}2-p_{i}-g_{i}}
\end{equation}
where $p_{i}$ and $g_{i}$ represent the confidence of $i$-th voxel grid on the reconstructed volume and ground truth.

\subsection{Implementation Details}
To train the R3D-SWIN,We use the AdamW\cite{ref17} optimizer with a learning rate of 1e-4,$\beta_{1}=0.9$,$\beta_{2}=0.999$,and a weight decay of 1e-2,epoch$=300$.The batch size is set to 16 for all the experiments.We use one A40 GPU in our experiments.Traning takes 2 days,depending on the exact setting.
\begin{table*}
\caption{QUANTITATIVE COMPARISON WITH OHTER SOTA METHODS ON SHAPENET,\textbf{BOLD} IS THE BEST PERFORMANCE.}
\label{table:1}
\begin{tabular}{cccccc}
\hline
Category   & Pix2vox++   & 3D-RETR     & UMIFormer    & Long-Range Grouping Transformer for Multi-View 3D Reconstruction & R3D-SWIN(Ours)     \\
\hline
airplane   & 0.673/0.549 & 0.705/0.593 & 0.685/       &                                                                  & \textbf{0.725/0.617} \\
bench      & 0.607/0.445 & 0.654/0.498 & 0.619/       &                                                                  & \textbf{0.679/0.518} \\
cabinet    & 0.798/0.405 & 0.808/0.422 & 0.798/       &                                                                  & \textbf{0.829/0.442} \\
car        & 0.857/0.541 & 0.859/0.548 & 0.848/       &                                                                  & \textbf{0.875/0.581} \\
chair      & 0.581/0.286 & 0.589/0.292 & 0.589/       &                                                                  & \textbf{0.627/0.325} \\
display    & 0.548/0.285 & 0.566/0.290 & \textbf{0.600}/       &                                                                  & 0.596/\textbf{0.322} \\
lamp       & 0.456/0.319 & 0.478/0.328 & 0.507/       &                                                                  & \textbf{0.507/0.360} \\
speaker    & 0.720/0.282 & 0.727/0.302 & 0.738/       &                                                                  & \textbf{0.757/0.321} \\
rifle      & 0.617/0.547 & 0.671/0.606 & 0.671/       &                                                                  & \textbf{0.679/0.615} \\
sofa       & 0.724/0.375 & 0.736/0.387 & 0.730/       &                                                                  & \textbf{0.762/0.421} \\
table      & 0.619/0.379 & 0.626/0.387 & 0.644/       &                                                                  & \textbf{0.659/0.415} \\
telephone  & 0.809/0.613 & 0.768/0.542 & 0.789/       &                                                                  & \textbf{0.816/0.609} \\
watercraft & 0.602/0.383 & 0.636/0.418 & 0.625/       &                                                                  & \textbf{0.659/0.451} \\

Overall    & 0.670/0.417 & 0.679/0.432 & 0.684/0.4281 & 0.6962 / 0.4461                                                  & \textbf{0.706/0.461} \\
\hline
\end{tabular}

\end{table*}

\section{EXPERIMENTS}
\label{sec:guidelines}
\begin{table*}
\caption{VISUALIZATION RESULTS ON SHAPENET.}
\label{table:2}
  \centering
  \begin{tabular}{  c  c  c  c c  c }
    \hline
    input & Pix2vox++ & 3D-RETR & UMIFormer & R3D-SWIN(Ours) & GT \\ \hline
    \begin{minipage}{.15\textwidth}
      \includegraphics[width=\linewidth]{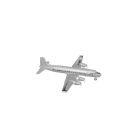}
    \end{minipage}
    &\begin{minipage}{.15\textwidth}
      \includegraphics[width=\linewidth]{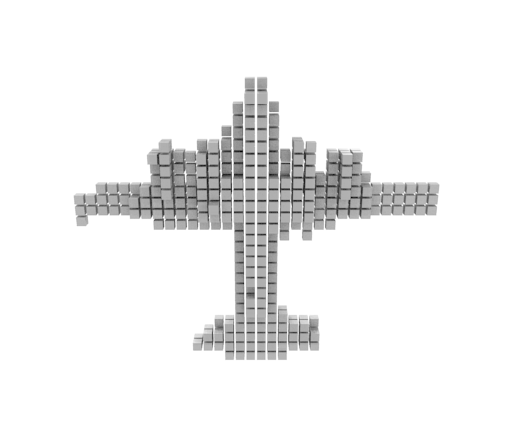}
    \end{minipage}
     &\begin{minipage}{.15\textwidth}
      \includegraphics[width=\linewidth]{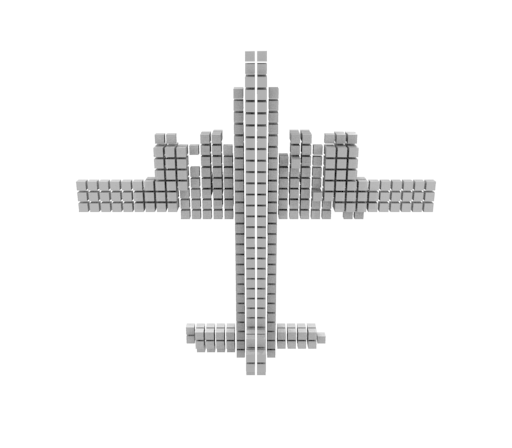}
    \end{minipage}
     &\begin{minipage}{.15\textwidth}
      \includegraphics[width=\linewidth]{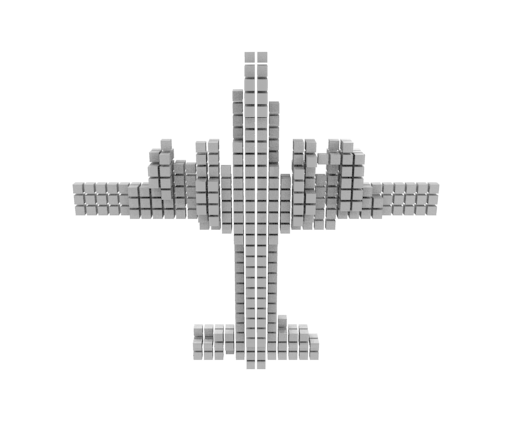}
    \end{minipage}
     &\begin{minipage}{.15\textwidth}
      \includegraphics[width=\linewidth]{shapenet_output/UMIFormer/UMIformerplane.png}
    \end{minipage}
    &\begin{minipage}{.15\textwidth}
      \includegraphics[width=\linewidth]{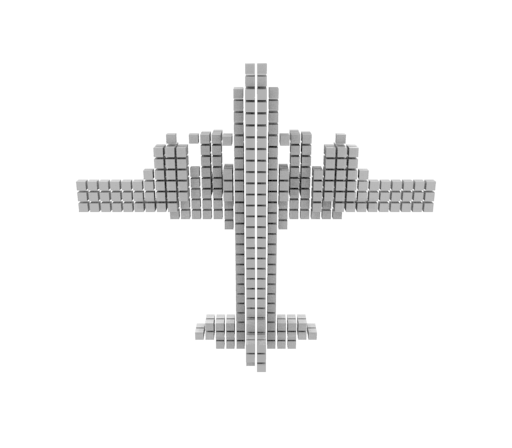}
    \end{minipage}\\

    \begin{minipage}{.15\textwidth}
      \includegraphics[width=\linewidth]{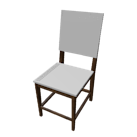}
    \end{minipage}
    &\begin{minipage}{.15\textwidth}
      \includegraphics[width=\linewidth]{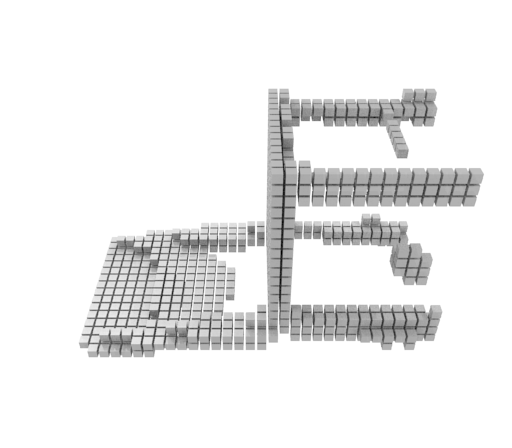}
    \end{minipage}
     &\begin{minipage}{.15\textwidth}
      \includegraphics[width=\linewidth]{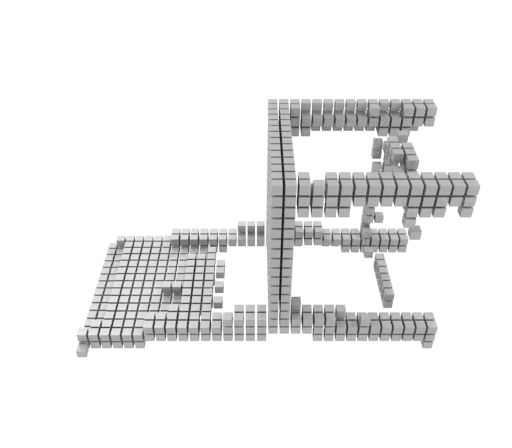}
    \end{minipage}
     &\begin{minipage}{.15\textwidth}
      \includegraphics[width=\linewidth]{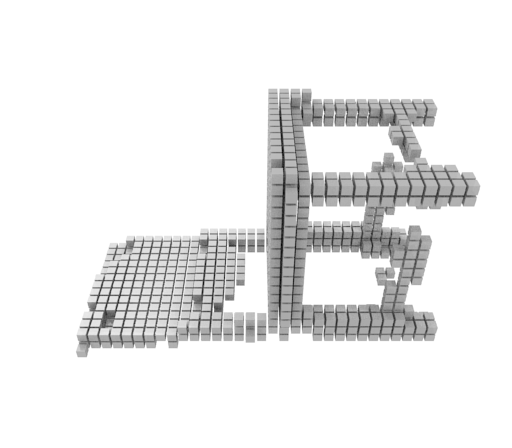}
    \end{minipage}
     &\begin{minipage}{.15\textwidth}
      \includegraphics[width=\linewidth]{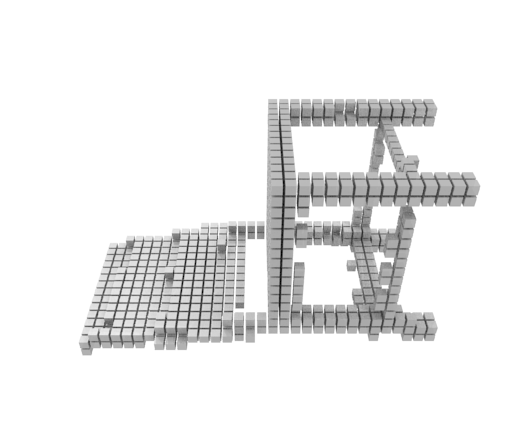}
    \end{minipage}
    &\begin{minipage}{.15\textwidth}
      \includegraphics[width=\linewidth]{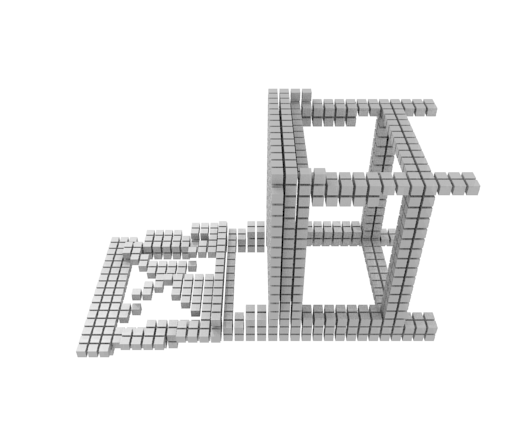}
    \end{minipage}\\

    \begin{minipage}{.15\textwidth}
      \includegraphics[width=\linewidth]{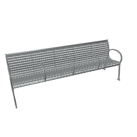}
    \end{minipage}
    &\begin{minipage}{.15\textwidth}
      \includegraphics[width=\linewidth]{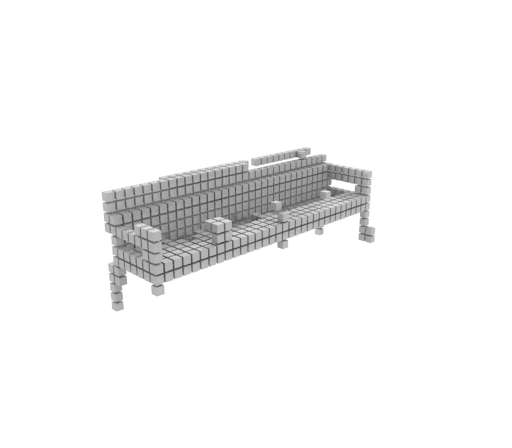}
    \end{minipage}
     &\begin{minipage}{.15\textwidth}
      \includegraphics[width=\linewidth]{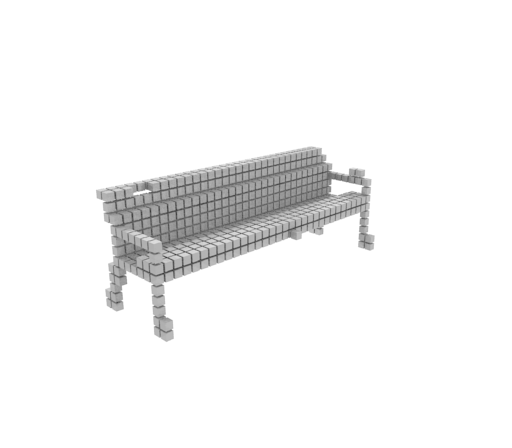}
    \end{minipage}
     &\begin{minipage}{.15\textwidth}
      \includegraphics[width=\linewidth]{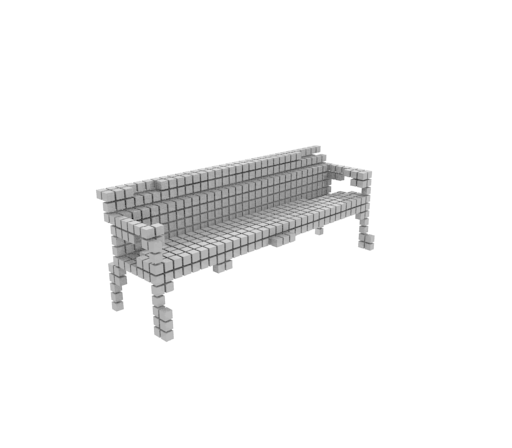}
    \end{minipage}
     &\begin{minipage}{.15\textwidth}
      \includegraphics[width=\linewidth]{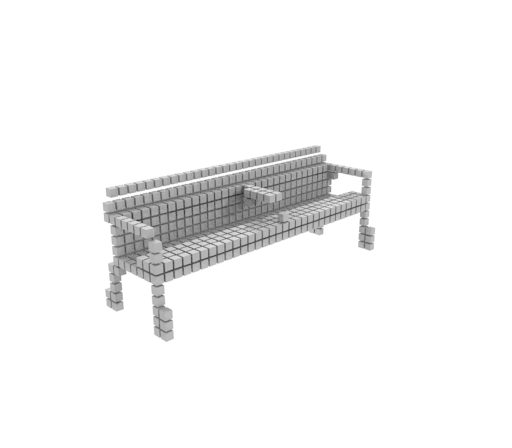}
    \end{minipage}
    &\begin{minipage}{.15\textwidth}
      \includegraphics[width=\linewidth]{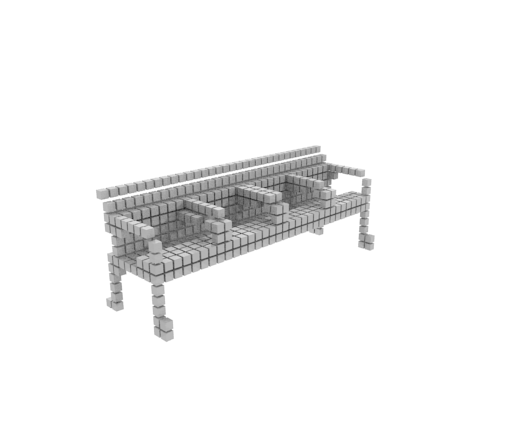}
    \end{minipage}\\

    \begin{minipage}{.15\textwidth}
      \includegraphics[width=\linewidth]{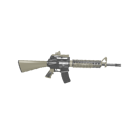}
    \end{minipage}
    &\begin{minipage}{.15\textwidth}
      \includegraphics[width=\linewidth]{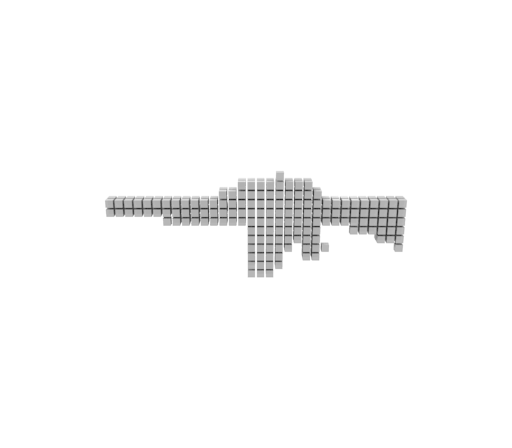}
    \end{minipage}
     &\begin{minipage}{.15\textwidth}
      \includegraphics[width=\linewidth]{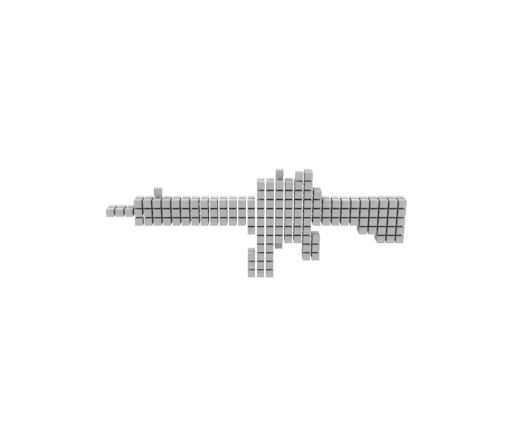}
    \end{minipage}
     &\begin{minipage}{.15\textwidth}
      \includegraphics[width=\linewidth]{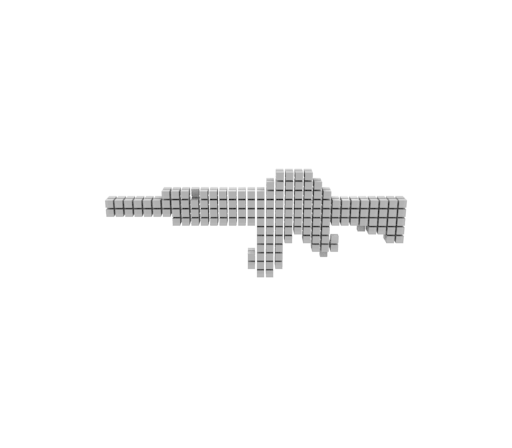}
    \end{minipage}
     &\begin{minipage}{.15\textwidth}
      \includegraphics[width=\linewidth]{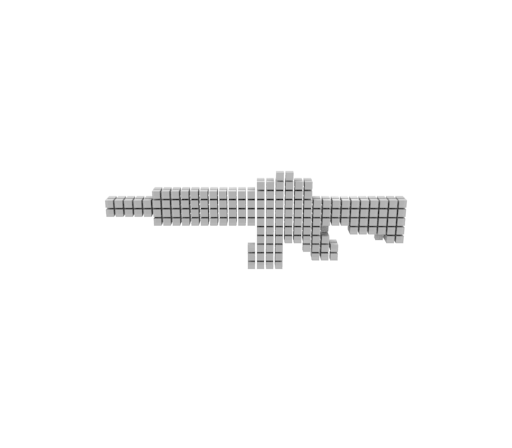}
    \end{minipage}
    &\begin{minipage}{.15\textwidth}
      \includegraphics[width=\linewidth]{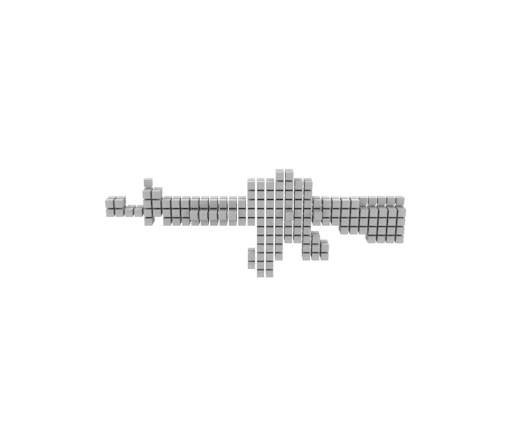}
    \end{minipage}

    \\ \hline
  \end{tabular}
\end{table*}

\begin{table*}
\caption{VISUALIZATION RESULTS ON SHAPENET.}
\label{table:3}
  \centering
  \begin{tabular}{  c  c    c c  c }
    \hline
    input & Pix2vox++  & UMIFormer & R3D-SWIN(Ours) & GT \\ \hline
    \begin{minipage}{.15\textwidth}
      \includegraphics[width=\linewidth]{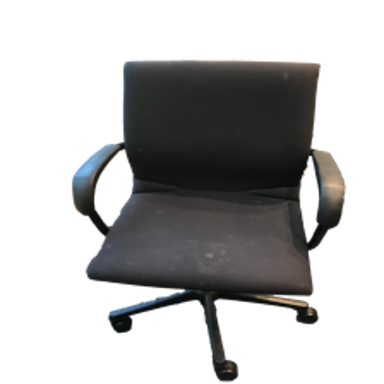}
    \end{minipage}
    &\begin{minipage}{.15\textwidth}
      \includegraphics[width=\linewidth]{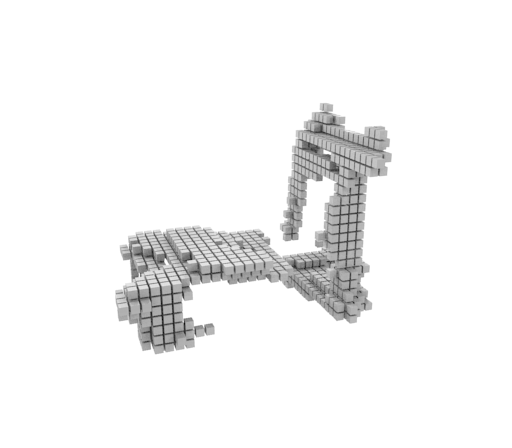}
    \end{minipage}
   
     &\begin{minipage}{.15\textwidth}
      \includegraphics[width=\linewidth]{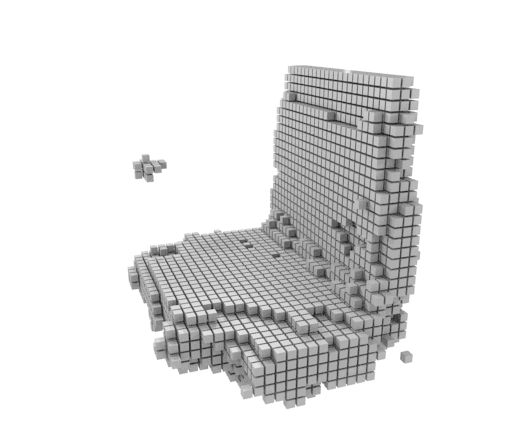}
    \end{minipage}
     &\begin{minipage}{.15\textwidth}
      \includegraphics[width=\linewidth]{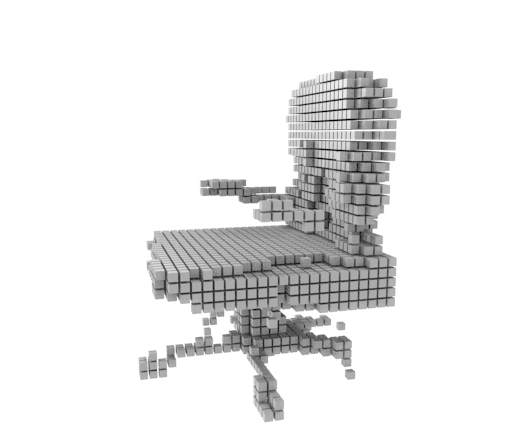}
    \end{minipage}
    &\begin{minipage}{.15\textwidth}
      \includegraphics[width=\linewidth]{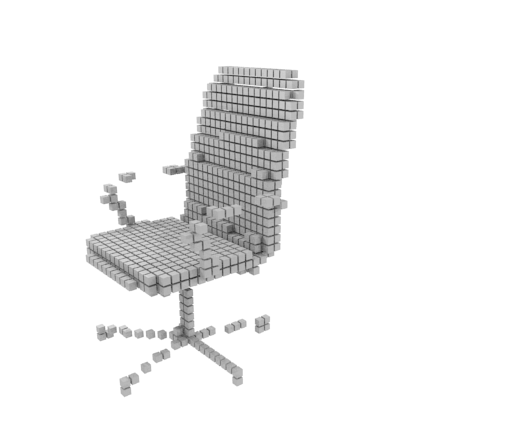}
    \end{minipage}\\

    \begin{minipage}{.15\textwidth}
      \includegraphics[width=\linewidth]{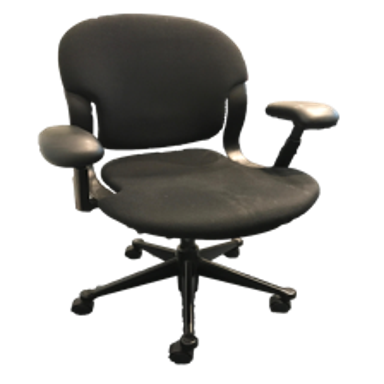}
    \end{minipage}
    &\begin{minipage}{.15\textwidth}
      \includegraphics[width=\linewidth]{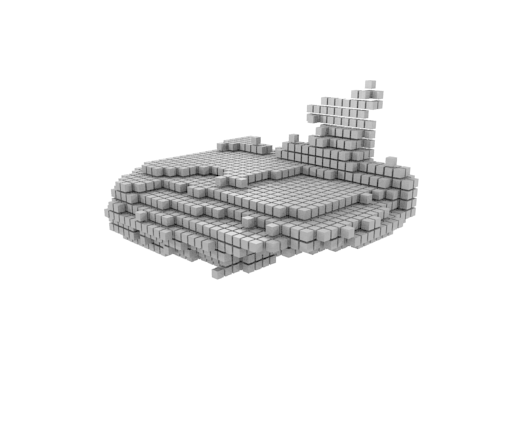}
    \end{minipage}
   
     &\begin{minipage}{.15\textwidth}
      \includegraphics[width=\linewidth]{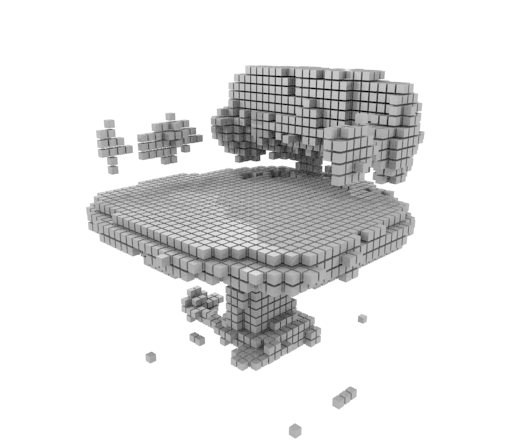}
    \end{minipage}
     &\begin{minipage}{.15\textwidth}
      \includegraphics[width=\linewidth]{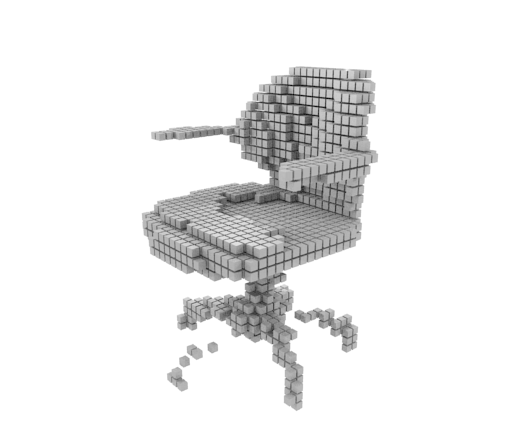}
    \end{minipage}
    &\begin{minipage}{.15\textwidth}
      \includegraphics[width=\linewidth]{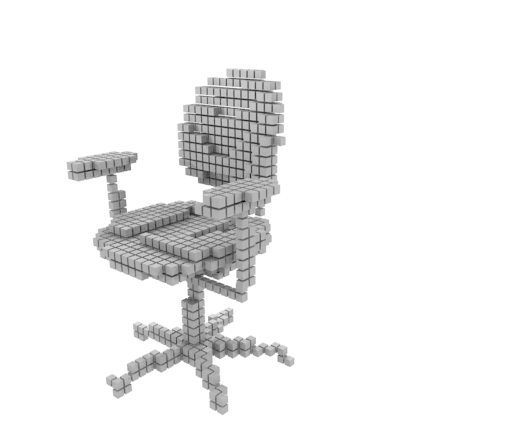}
    \end{minipage}\\

    \\ \hline
  \end{tabular}
\end{table*}

\subsection{Evaluation metric}
3D Intersection over Union (IoU) and F-score@\%1 are used to measure the reconstruction performance.Higher IoU and Fscore values indicate better performance.

The mean Intersection-over-Union (IoU) is formulated as:
\begin{equation}
IoU=\frac{\sum _{\left ( i,j,k \right )}I\left ( \hat{p}\left ( i,j,k \right )> t \right )I\left ( p\left ( i,j,k \right ) \right )}{\sum _{\left ( i,j,k \right )}I\left [  I\left ( \hat{p}\left ( i,j,k \right )> t \right )+I\left ( p\left ( i,j,k \right ) \right )\right ]}             
\end{equation}
where $ \hat{p}\left ( i,j,k \right )$ and $ p\left ( i,j,k \right )$ represent the predicted occupancy probability and ground truth at $(i, j, k)$. $I(·)$ is an indicator function and t denotes a voxelization threshold.

The F-score@\%1 is formulated as:
\begin{equation}
Fscore\left ( d \right )=\frac{2P\left ( d \right )R\left ( d \right )}{P\left ( d \right )+R\left ( d \right )}    
\end{equation}
where $P\left ( d \right )$ and $R\left ( d \right )$ denote the precision and recall for a distance threshold between prediction and ground truth.F-Score@1\% indicates the F-Score value when $d$ is set to $1\%$.
\subsection{Datasets}
ShapeNet\cite{ref14} is a large-scale 3D object dataset consisting of 55 object categories with 51,300 3D models. Following the setting in Pix2Vox\cite{ref3}, we use the same subset of 13 categories and about 44,000 models.

Different from ShapeNet\cite{ref14}, in which all examples are synthetic, Pix3D\cite{ref15} is a dataset of aligned 3D models and real-world 2D images. Evaluating models on Pix3D gives a better understanding of the model performance under practical settings.

\subsection{Results}\label{formats}
For single-view 3D reconstruction on ShapeNet\cite{ref14}, we compare our results with previous state-of-the-art models, including Pix2vox++\cite{ref4},3dretr\cite{ref8},Umiformer\cite{ref9},\cite{ref10}.Table \ref{table:1} shows the results,table \ref{table:2} show visualization results.We can observe that our method outperform all previous models in terms of overall IoU and F1 score. Additionally, our method outperforms all other baselines in 12 of the 13 categories, except the Iou of display.

Following the same setting in Pix3D\cite{ref15}, we use the subset consisting of 2,894 untruncated and unoccluded chair images as the test set. Moreover, we follow Legoformer\cite{ref7} to  use the mask to segment out the background and replace it with
a constant color. The resulting images are provided as input to R3D-SWIN trained only on Shapenet and we report examples of the predicted models  in table\ref{table:3}.

\begin{table*}
\caption{comparisons of parameter size and model performance}
\label{table:4}
\centering
\begin{tabular}{lll}
\hline
Pretrained model &Param  &Iou overall \\
\hline
swin tiny      &29.3M     &0.687   \\
swin small     & 50.7M    &0.689   \\
swin base      & 88.8M    &0.690  \\
swinv2 base    &89.1M     &\textbf{0.706}         \\
\hline
\end{tabular}
\end{table*}

\begin{table*}
\caption{diffferent version of Swin Transformer encoder performerance and accuracy}
\label{table:5}
\centering
\begin{tabular}{llllll}
\hline
module       & Pix2vox++ & 3D-RETR          & UMIFormer        & R3DSWIN(ours)   & R3D-SWIN*        \\
\hline
encoder      & 5.5M      & 85.8M            & 88.1M            & 87.9M            & 87.9M            \\
decoder      & 55.8M     & 77.3M            & 76.9M            & 1.2M             & 76.7M            \\
refiner      & 34.8M     & \textbackslash{} & \textbackslash{} & \textbackslash{} & \textbackslash{} \\
merger       & 0.017M    & \textbackslash{} & 14.1M            & \textbackslash{} & \textbackslash{} \\
sum          & 96.1M     & 163.1M           & 179.1M            &\textbf{89.1M}       & 164.6M           \\
Iou\_overall & 0.67      & 0.679            & 0.684            & \textbf{0.706}            &0.686 \\
\hline
\end{tabular}
\end{table*}

\subsection{Ablation study}
Pretraining for the Swin Transformer Encoder is crucial since Transformers large amounts of data to gain prior knowledge for images. In this setup, we observe that the performance of R3D-SWIN decreases significantly without pretraining.We show diffferent version of Swin Transformer encoder performerance and accuracy in table\ref{table:4}.

We give further comparisons of parameter size and model performance in Table\ref{table:5}. Despite that our R3D-SWIN is smaller than previous state-of-the-art models, it still reaches
better performance.To verify the effectiveness of our encoder, we compared it with the original 3dretr\cite{ref8} encoder.R3D-SWIN* use 3dretr\cite{ref8} decoder directly.

\section{Conclusion and Limitations}
In this paper, we propose a novel transformer-based network for single-view 3D reconstruction which achieves SOTA accuracy.The encoder uses moving window attention to extract image features from single-view data, and a simple decoder converts image features into voxel models.The major limitation of our proposed method is that it does not achieve sota accuracy in multi-view performance.We will study multi-view 3D reconstruction in future work.

\end{document}